# Rubik's Cube Operator: A Plug And Play Permutation Module for Better Arranging High Dimensional Industrial Data in Deep Convolutional Processes

Luoxiao Yang, Zhong Zheng, and Zijun Zhang, *Senior Member, IEEE*

**Abstract**—The convolutional neural network (CNN) has been widely applied to process the industrial data based tensor input, which integrates data records of distributed industrial systems from the spatial, temporal, and system dynamics aspects. However, unlike images, information in the industrial data based tensor is not necessarily spatially ordered. Thus, directly applying CNN is ineffective. To tackle such issue, we propose a plug and play module, the Rubik's Cube Operator (RCO), to adaptively permutate the data organization of the industrial data based tensor to an optimal/suboptimal order of attributes before being processed by CNNs, which can be updated with subsequent CNNs together via the gradient-based optimizer. The proposed RCO maintains K binary and right stochastic permutation matrices to permutate attributes of K axes of the input industrial data based tensor. A novel learning process is proposed to enable learning permutation matrices from data, where the Gumbel-Softmax is employed to reparameterize elements of permutation matrices, and the soft regularization loss is proposed and added to the task-specific loss to ensure the feature diversity of the permuted data. We verify the effectiveness of the proposed RCO via considering two representative learning tasks processing industrial data via CNNs, the wind power prediction (WPP) and the wind speed prediction (WSP) from the renewable energy domain. Computational experiments are conducted based on four datasets collected from different wind farms and the results demonstrate that the proposed RCO can improve the performance of CNN based networks significantly.

—————————— ◆ ——————————

## 1 INTRODUCTION

RECENT advancement of sensing and data storage technologies, such as the supervisory control and data acquisition (SCADA) system, in the industrial sector has enabled an unprecedented opportunity of collecting the massive volume of data from distributed industrial systems and one widely studied representative is the distributed renewable energy systems, such as wind turbines and photovoltaic systems [5-15]. Inspired by successes of deep learning in computer vision and natural language processing applications [3], latest studies of data-driven renewable energy system modeling methods have largely converged to the deep learning based paradigms [5-15]. One apparent advantage of adopting deep learning in renewable energy systems is that a much higher dimension of input data can be considered in data-driven modeling problems, which integrate data records of distributed systems from the spatial, temporal, and system dynamics aspects as a tensor structure. To process such tensor input of an extreme high dimension, the convolutional neural network (CNN) [1] dominating the computer vision field was naturally adopted in [5-15] due to its advantageous properties including the sparse connectivity, parameter sharing, subsampling, and local receptive fields [2]. The convolutional operation is originally designed for processing images, a tensor of pixels spatially ordered. Yet, orders of attributes in certain dimensions of industrial data based tensor inputs are exchangeable, which means that elements in such tensor is not necessarily ordered. As the kernel based feature extraction in convolutional operation can be approximately considered as a local regressor, orders of attributes can lead to various outcomes of data feature engineering. Therefore, to obtain a more effective feature engineering of such tensor input via CNNs, one grand challenge needs to be tackled is attaining the optimal order of attributes in such tensor for the convolutional process.

Such critical issue has not been studied in the literature and most studies mainly focused on directly applying or adapting CNN and its variants to uplift the application performance [5-16]. Next, we further elaborate the inefficacy of directly adopting CNNs for processing industrial data based tensor inputs and the necessity of designing a new mechanism in details. Fig. 1 illustrates the difference between an image and an industrial data based tensor visually. It is clear that pixels in the image are spatially ordered to present the context. Yet, the industrial data

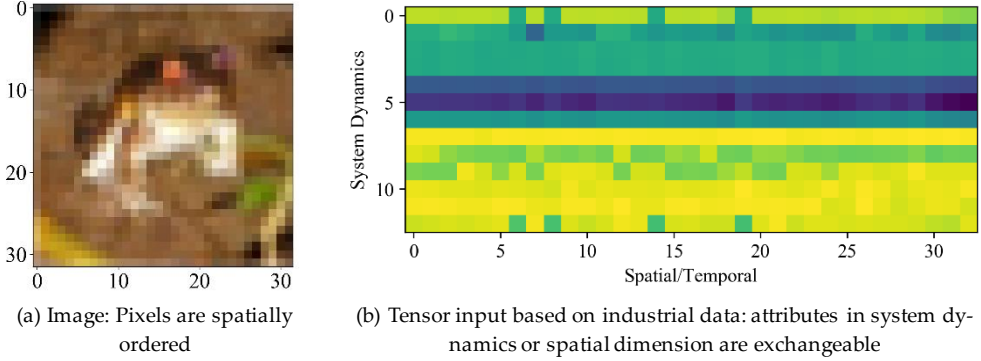

(a) Image: Pixels are spatially ordered

(b) Tensor input based on industrial data: attributes in system dynamics or spatial dimension are exchangeable

Fig. 1. An illustration of the image-structured data and the non-image-structured industrial data.

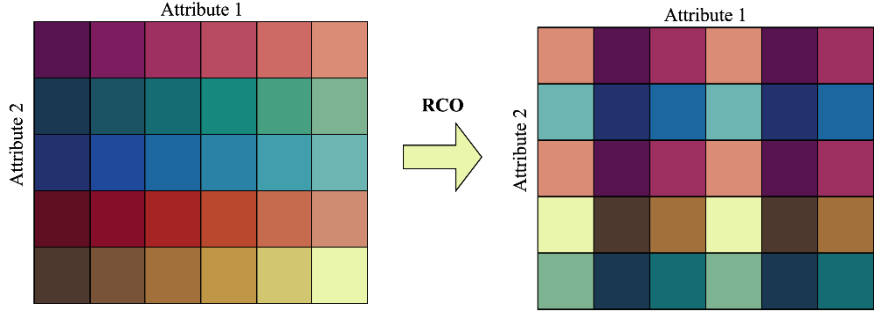

Fig. 2. An illustration of the non-image-structured industrial data before and after Rubik's Permutation.

based tensor is not bounded by a visual context. Only attributes along the temporal dimension are ordered while orders of attributes in other dimensions are arbitrary and their exchanges do not affect the total conveyed information. The generic CNN designed for processing images is well known with two main properties, the local correlation and the shift invariance [3]. As the industrial data based tensor is not necessarily ordered, different orders of data arrangement will obviously return different local correlations among data via convolutions in the downstream feature engineering and modeling tasks. Moreover, as different attributes in the dimension of system dynamics and spatial-temporal locations have their physical meaning, such industrial data based tensor is not shift invariant. Due to the previously described unique characteristics of the industrial data based tensor, directly applying the CNN into processing such tensor might diminish the superior properties of the CNN. In light of this, to process such industrial data based tensor, a new convolutional process adaptively permutating the data organization in such tensor to an optimal order of attributes arrangement before convolutional operations needs to be studied.

In this paper, we propose a plug and play module, the Rubik's Cube Operator (RCO), to enable an automated arrangement of orders of attributes in processing industrial data based tensor inputs via CNNs. Given a $K$-dimensional tensor, the proposed RCO maintains $K$ permutation matrices to permute attribute orders on $K$ axes of such tensor and next feeds the permuted data into the CNN for different downstream learning tasks. To develop such RCO, we need to address a critical challenge, a method for efficiently obtaining suitable permutation matrices, which is currently unavailable. We develop a unique learning process, which allows permutation matrices to be trained together with the CNN via a gradient-descent and backpropagation process based on data. Permutation matrices are not directly learnable as by definition, they are square binary matrices. We introduce the Gumbel-Softmax [4] to reparameterize elements in them to enable the pass of gradients while ensuring that learnt outcomes are binary and right stochastic. Meanwhile, a soft regularization loss is proposed and added to the task-specific loss to ensure the feature diversity of the permuted data. Fig. 2 visualizes the industrial data based tensor before and after RCO. It is observable that the permutation operation reshapes the original data, and results in a few repeated local correlations. It looks like that data organization is permuted to present some local correlations and shift invariances, which are beneficial to following feature engineering processes via the CNN. To illustrate the practical value of RCO, we consider two recently famous learning tasks in the renewable energy field, developing the CNN based models for the wind power prediction (WPP) and the wind speed prediction (WSP) with considering tensor inputs incorporating data from the spatial, temporal, and system dynamics aspects. Computational experiments are conducted to develop CNN models with and without RCOs for two tasks to validate the value of the proposed RCO. Experimental results demonstrate that with the RCO module, the performance of both learning tasks considered is improved significantly.

Main contributions of this paper are listed as follows:
1. A novel plug and play RCO module is introduced to better accommodate the industrial data based tensor in the CNN based data processing for the first time.
2. A theoretical analysis explaining the effectiveness and several properties of the proposed RCO is presented.
3. The proposed RCO can facilitate CNNs to achieve better performance on processing the industrial data based tensor input, which is justified via a careful computational study considering representive real industrial problems and data.

## 2 RELATED WORKS

We would like to first re-visit two groups of literature relevant to our work, the recent CNN based industrial data processing and the permutation learning.

### 2.1 The CNN Based Industrial Data Processing

Recently, a great interest in industrial distributed system data analytics has switched from the traditional tabular-form data to a tensor-form data which is of much larger dimensions. This is largely motivated by the natural expansion of the data-driven modeling study focus from a single unit to the spatial temporal impacts from its neighboring units. To enable processing the tensor-form data and extract valuable data features, the CNN has been widely adopted in various industrial applications covering the renewable energy prediction [5-12], load forecasting [13, 14], anomaly detection [15, 16], transportation accident analysis [17], etc.

In [5], a CNN-based method was developed to capture the spacial relations of neighboring wind farms in the short-term wind speed prediction. In [6], wind speed data of a wind farm was re-arranged into a 3-dimensional tensor and the CNN-long short-term memory unit (LSTM) was applied to capture patterns of the spatial-temporal correlation. In [7], a 12-hidden-layer deep CNN was employed to capture the wind speed spatial correlation and the output latent representation was fed into a LSTM for the final forecasting. In [8], a deep attention CNN-LSTM based model was proposed for short-term wind speed prediction, where the CNN was utilized to extract the within-site correlation. In [9], a unique CNN-Gated Recurrent Units (GRU) based model with a two-layer clus-tersing based data scenario recognition was designed for the wind power prediction. In [10], a hybrid CNN-LSTM model was proposed for the household load forecasting, where the 1-dimensional CNN was employed to capture the local tendency. In [11-12], a hybrid CNN-LSTM model was also presented to predict the PV plant power production, while the CNN was applied to focus on extracting correlations between the physics-based and spatial dimension. In [13], a CNN integrated with a support vector regression (SVR) was proposed for the load forecasting, and the CNN mainly served extracting the temporal relationship between the temperature and load components. In [14], the CNN-LSTM concept was applied into the short-term load forecasting and the CNN was mainly used to capture data temporal patterns. In [15], a conditional convolutional autoencoder was developed for the wind turbine (WT) blade breakage detection task. The CNN was considered as the backbone for developing the convolutional autoencoder of processing the SCADA data. In [16], a LSTM-based CNN was designed for the anomaly detection of autonomous vehicles and the CNN served processing data from multiple sensors. In [17], a novel traffic accident severity prediction-convolutional neural network (TASP-CNN) model was developed to predict the severity of traffic accidents and the CNN was employed to engineer features meaningful for identifying traffic accidents. All aforementioned studies have reported the superior performance of CNNs on capturing patterns, such as the spatial-temporal correlation, in the industrial data based tensor input. However, orders of attributes on each axis of the tensor input are usually arbitrarily determined and investigations on an optimal order of arranging attributes in the tensor input for the CNN based data processing have not been conducted.

### 2.2 Permutation Learning

Permutation learning aims to revert disordered fragments of a data, such as an image, to their original order via permutation matrices. It has been discussed in applications including the computer vision [18], robotics [19], ranking [20], etc.

The generic permutation learning problem aims to learn a square doubly stochastic permutation matrix $P \in R^{n \times n}$ to restore the order of disordered fragments of a data $X \in R^n$ via a matrix multiplication operation. The objective function of the generic permutation learning is expressed in (1):

$$\min_P \|X_t - PX\|$$
$$s.t. \quad P = (P_{ij}) \in \{0,1\}^{n \times n}$$
$$\sum_i P_{ij} = 1 \quad i = 1,2,\dots,n \quad (1)$$
$$\sum_j P_{ij} = 1 \quad j = 1,2,\dots,n$$

where $X_t$ denotes the ground truth. Directly learning $P$ is intractable due to its constraints and binary elements. To enable learning $P$ via data, in the literature [18-19], it is first relaxed to a matrix of real elements between [0, 1] via the Gumbel-Softmax and the Sinkhorn operation, which, after convergence, ensure that constraints in (1) can be satisfied. The permutation learning has achieved a great success in many applications. In [18], an end-to-end neural network (NN), Sinkhorn network, was proposed to learn the permutation matrix. It was composed of three steps. First, a deep NN was utilized to extract the latent representation of disordered fragments of a targeted data and output the permutation matrix. Next, the continuous Sinkhorn operator and the Gumbel-Softmax were adopted to iteratively update the produced permutation matrix to make it binary and doubly stochastic. Finally, the Sinkhorn networks were updated via backpropagation based on an end-to-end learning. In [19], a Sinkhorn networks based method was presented for learning the sequence of robot actions. In [21], a Lehmer Codes-based method was developed to learn the permutation matrix. The Lehmer Codes mapped permutations into a Euclidean vector space such that the distance between the prediction and the ground truth could be easily measured.

Intuitively, the idea of permutation learning could be a good fit to obtain appropriate arrangements of attributes on axes of industrial data based tensor inputs. However, the generic permutation learning cannot be simply adapted into such task due to following three factors. First, there does not exist a ground truth permutation for the industrial data based tensor input because different tasks require different local features. Therefore, the optimal permutation matrices can be significantly different over different tasks. Secondly, addressing each single task itself is time-consuming and needs a huge amount of computational resources. Sinkhorn networks [18-19] require repeated iterations, which could significantly increase the computational burden. Finally, existing permutation methods assume that the permutation matrix allows each feature to appear only one time, which might not be such strict in industrial data. In this case, the shift invariance propriety of the permuted data cannot be accommodated.

To host unique needs of processing the industrial data based tensor input, we would like to propose a new mechanism for learning permutation matrices, the Rubik's Cube Operator, to permutate the information on axes of such tensor so that the subsequent feature engineering via the CNN and the final learning task can be benefited.

## 3 THE RUBIK'S CUBE OPERATOR

To introduce the proposed plug and play RCO module, we first formulate the problem of learning such RCO. Next, we detail the learning process and preliminarily explore its properties from the theoretical aspect.

### 3.1 The problem formulation

Denote an industrial data based tensor input of manually ordered attributes as $X_0 \in R^{n_1 \times n_2 \times \ldots \times n_K}$. The objective function of a learning task which applies CNNs to process the industrial data based tensor input is expressed in (2):

$$\min_{\theta} \mathcal{L}_T(f(X_0), label) \quad (2)$$

where $\mathcal{L}_T$ denotes the task-specific loss function, $f(\cdot)$ denotes the CNN-based network and $\theta$ denotes the trainable parameters of $f(\cdot)$. The *label* denotes the ground-truths of the task.

The RCO is proposed to permutate the order of attributes on $K$ axes of $X_0$ in (2) via introducing $K$ permutation matrices, just like switching a rubik's cube, so that local correlation and shift invariance properties in $X_0$ can be more prominent. Such permutation is described in (3):

$$X_k = SwapAxes(P_k * SwapAxes(X_{k-1}, 1, k), 1, k)$$
$$k = 1, 2, \ldots, K \quad (3)$$

where $SwapAxes(X, 1, k)$ denotes the operation interchanging axes 1 and $k$ of the tensor $X$ and the $P_k$ denotes the permutation matrix for the axis $k$. There does not exist a ground truth of the order of elements in the industrial data based tesor input. The objective of learning the RCO is thus connected with the downstream learning task, which aims to offer an optimal permutation on $X_0$ so that the performance of downstream tasks can be improved, as described in (4):

$$\min_{P_1, P_2, \ldots, P_K, \theta} \mathcal{L}_T(f(X_K), label)$$
$$s.t. \quad P_k = \left(P_{k_{ij}}\right) \in \{0,1\}^{n_k \times n_k} \quad k = 1, 2, \ldots, K \quad (4)$$
$$\sum_j P_{k_{ij}} = 1 \quad k, j = 1, 2, \ldots, K$$

For convenience, we call the first constraint of (4), $P_k = \left(P_{k_{ij}}\right) \in \{0,1\}^{n_k \times n_k} \quad k = 1, 2, \ldots, K$, as the **binary condition** and the second constraint, $\sum_j P_{k_{ij}} = 1 \quad k, j = 1, 2, \ldots, K$, as the **right stochastic condition**.

The main difference between the generic permutation learning and the RCO learning for processing the industrial data based tensor is of two-fold. First, the objective function of two learning problems is different. The generic permutation learning aims to restore the disordered fragments of images, the disordered sequential actions, etc. Thus, there always exists a ground truth order. However, the RCO learning aims to better organize the industrial data based tensor and the optimal permutation is intractable. Secondly, the traditional permutation matrix is a square doubly stochastic matrix, while the proposed RCO Permutation matrix is a square right stochastic matrix. This setting is designed based on unique characteristics of processing industrial data based tensor inputs. The generic permutation learning mainly focuses on restoring the order of data pieces and each piece is allowed to appear only once. In contrasts, the RCO learning for re-organizing the industrial data based tensor aims to enhance its local correlation and shift invariance to facilitate the latter convolutional process, both of which allow important features to appear more than one time.

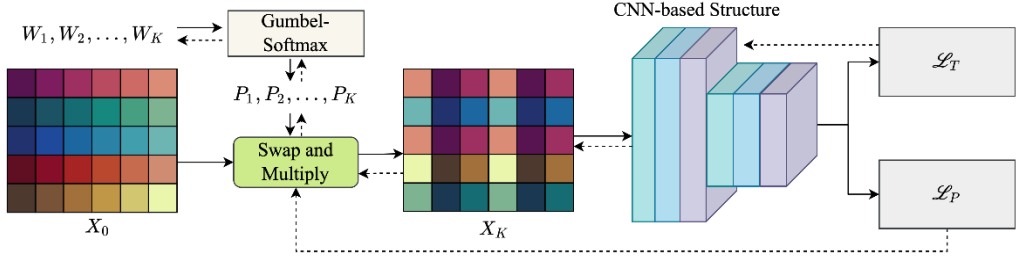

Fig. 4. The learning process of the proposed Rubik's Cube Operator and the Subsequent CNN-based structure.

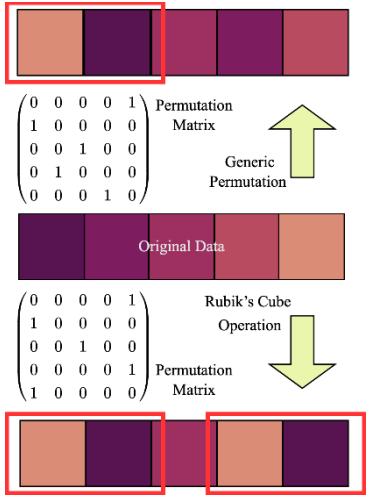

Fig. 3. An illustration of the generic permutation and the Rubik's Cube Operation.

Fig. 3 visually compares the generic permutation and the Rubik's Cube Operation, where the red rectangle box presents the significant locally correlated features. We clearly observe following points. First, by introducing the permutation matrix, both the generic permutation and the Rubik's Cube Operation can rearrange the original disordered data and enhance the local correlations. Secondly, by changing the doubly stochastic matrix into the right stochastic matrix, the RCO allows significant locally correlated features to appear repeatedly. Thus, the shift invariant property of the rearranged industrial data based tensor can be enhanced by the Rubik's Cube operation. Next, we will introduce details of the process for learning the RCO described in (4).

### 3.2 The learning process

The learning of the proposed RCO is conducted together with the subsequent CNN-based structure in a specific data-driven modeling task. The overall learning process of the proposed RCO and connected CNN-based structure is presented Fig. 4. The dashed lines in Fig. 4 indicate the gradient flow of the permutation matrices in the RCO. First, a set of trainable parameters $W_1 \in R^{n_1 \times n_1}, W_2 \in R^{n_2 \times n_2}, \ldots, W_K \in R^{n_K \times n_K}$ reparameterizing elements of permutation matrices are initialized by filling with ones. After the initialization, all permutation matrices, $P_1, P_2, \ldots, P_K$ are computed based on the Gumbel-Softmax to satisfy the binary condition and the right stochastic condition as described in (5):

$$P_{k_{ij}} = \frac{e^{\frac{W_{k_{ij}}}{\tau}}}{\sum_{u=1}^{n_k} e^{\frac{W_{k_{iu}}}{\tau}}} \quad (5)$$

where $\tau$ denotes the temperature hyperparameter that will decline during the training. Next, $X_k$ is computed based on (3) and fed into a predefined CNN-based structure. The elements of the permutation matrix, $W_1, W_2, \ldots, W_K$, will be updated according to the loss function $\mathcal{L}_T$. To avoid the loss of feature diversity that the permutation matrix $P_k$ may only pay attention to the most important feature such that $\sum_{i=1}^{n_k} P_{k_{ij}} = n_k$ and then result in overfitting, we design the soft regularization loss $\mathcal{L}_P$ as expressed in (6):

$$\mathcal{L}_P(P_1, P_2, \ldots, P_K) = \sum_{k=1}^{K} ReLU\left(\frac{1}{n_k}\sum_{j=1}^{n_k}\left(\sum_{i=1}^{n_k} P_{k_{ij}} - 1\right)^2 - \gamma * n_k\right) \quad (6)$$

where $\gamma \in [0, 1]$ denotes the degree of the soft regularization loss. Suppose that the binary condition and the right stochastic condition are satisfied. By setting $\gamma = 1$, $\mathcal{L}_P$ does not restrict the feature diversity. In contrast, if $\gamma = 0$, $\mathcal{L}_P$ forces $P_k$ to be a doubly stochastic

**Algorithm 1:** Learning process of the proposed RCO

1: **Input**: $X_0$, CNN-based structure $f(\cdot), \lambda, \gamma, \mathcal{L}_T, label$.
2: Initialize $W_1, W_2, ..., W_K$ and $\tau = 1$
3: **for** epoch $i=1$ to $N$
4: $P_1, P_2, ..., P_K = GumbelSoftmax(W_1, W_2, ..., W_K)$
5: $X_K = SwapAndMultiple(X_0, P_1, P_2, ..., P_K)$
6: Compute the loss function based on (7) and update $W_1, W_2, ..., W_K, \theta$ using the gradient-based optimizer
7: $\tau = \tau * 0.9$
10: **end for**

matrix, which means all features are utilized. Therefore, the final loss function of the proposed RCO learning is expressed in (7):

$$\mathcal{L} = \mathcal{L}_T(f(X_K), label) + \lambda * \mathcal{L}_P(P_1, P_2, ..., P_K) \quad (7)$$

where $\lambda$ denotes the hyperparameter to balance $\mathcal{L}_T$ and $\mathcal{L}_P$.

The Pseudocode of the proposed RCO learning process is presented in **Algorithm 1**.

We then examine from the theoretical aspect that, by adopting the Gumbel-Softmax, whether the permutation matrix $P_k$ could satisfy the binary condition and the right stochastic condition, and by setting $\gamma = 0$, whether the permutation matrix $P_k$ could converge to a doubly stochastic matrix.

**Assumption 1**. We assume that after $e$ epoch ($e \geq 1$), $\exists z \in \{1,2,...,n_K\}$ that (8) holds:

$$W_{k_{iz}} > W_{k_{ij}} \quad j = 1,2,...,z-1, z+1, ..., n_k \quad (8)$$

**Proposition 1**. *After sufficient training epochs, $P_k$ could meet the binary condition and the right stochastic condition.*

**Proof**. Based on the definition of the Gumbel-Softmax in (5), we have (9):

$$\sum_{j=1}^{n_k} P_{kij} = \sum_{j=1}^{n_k} \frac{e^{\frac{W_{kij}}{\tau}}}{\sum_{u=1}^{n_k} e^{\frac{W_{kiu}}{\tau}}} = 1 \quad (9)$$

With **Assumption 1**, we can obtain (10):

$$\lim_{\tau \to 0} P_{kij}|_{j=z} = \lim_{\tau \to 0} \frac{e^{\frac{W_{kiz}}{\tau}}}{\sum_{u=1}^{n_k} e^{\frac{W_{kiu}}{\tau}}} = \lim_{\tau \to 0} \frac{1}{\sum_{u=1}^{n_k} e^{\frac{W_{kiu} - W_{kiz}}{\tau}}}$$

$$= \lim_{\tau \to 0} \frac{1}{\sum_{u=1, u \neq z}^{n_k} e^{\frac{W_{kiu} - W_{kiz}}{\tau}} + 1} = 1$$

$$\lim_{\tau \to 0} P_{kij}|_{j \neq z} = \lim_{\tau \to 0} \frac{e^{\frac{W_{kij}}{\tau}}}{\sum_{u=1}^{n_k} e^{\frac{W_{kiu}}{\tau}}} = \lim_{\tau \to 0} \frac{e^{\frac{W_{kij} - W_{kiz}}{\tau}}}{\sum_{u=1}^{n_k} e^{\frac{W_{kiu} - W_{kiz}}{\tau}}}$$

$$= \lim_{\tau \to 0} \frac{e^{\frac{W_{kij} - W_{kiz}}{\tau}}}{\sum_{u=1, u \neq z}^{n_k} e^{\frac{W_{kiu} - W_{kiz}}{\tau}} + 1} = 0 \quad (10)$$

Thus, by adopting the Gumbel-Softmax, $P_k$ satisfies the binary condition and the right stochastic condition. This establishes **Proposition 1**.

**Proposition 2**. *If $\gamma = 0$, $P_k$ will converge to a doubly stochastic matrix with the appropriately chosen $\lambda$.*

**Proof**: With **Proposition 1**, after sufficient epochs, $P_k$ satisfies the binary condition and the right stochastic condition. Suppose $P_k$ is not a doubly stochastic matrix and (11) always holds:

$$\exists j_1, j_2, i_1, \in [1, n_k]$$
$$s.t. \sum_i P_{k_{ij_1}} > 1 > \sum_i P_{k_{ij_2}} \quad (11)$$
$$P_{k_{i_1 j_1}} = 1, \quad P_{k_{i_1 j_2}} = 0, \quad W_{k_{i_1 j_1}} > W_{k_{i_1 j_2}}$$

If $\gamma = 0$, the loss function could be expressed as (12):

$$\mathcal{L} = \mathcal{L}_T(f(X_K), label) + \lambda * \sum_{k=1}^{K} \frac{1}{n_k} \sum_{j=1}^{n_k} \left( \sum_{i=1}^{n_k} P_{kij} - 1 \right)^2 \quad (12)$$

The corresponding gradients of $W_{k_{i_1 j_1}}$ and $W_{k_{i_1 j_2}}$ are computed in (13):

$$\Delta W_{k_{i_1 j_1}} = \frac{\partial \mathcal{L}_T}{\partial W_{k_{i_1 j_1}}} + \frac{2\lambda}{n_k} \left( \sum_{i=1}^{n_k} P_{kij_1} - 1 \right) \frac{\partial P_{k_{i_1 j_1}}}{\partial W_{k_{i_1 j_1}}}$$

$$\Delta W_{k_{i_1 j_2}} = \frac{\partial \mathcal{L}_T}{\partial W_{k_{i_1 j_2}}} + \frac{2\lambda}{n_k} \left( \sum_{i=1}^{n_k} P_{kij_2} - 1 \right) \frac{\partial P_{k_{i_1 j_2}}}{\partial W_{k_{i_1 j_2}}} \quad (13)$$

$$\frac{\partial P_{kij}}{\partial W_{kij}} = \frac{1}{\tau} \frac{e^{\frac{W_{kij}}{\tau}} \sum_{u=1, u \neq j}^{n_k} e^{\frac{W_{kiu}}{\tau}}}{\left( \sum_{u=1}^{n_k} e^{\frac{W_{kiu}}{\tau}} \right)^2} > 0$$

We denote $\frac{\partial P_{kij}}{\partial W_{kij}}$ as $\Delta_{kij}$ for simplification. Considering the gradient-based optimizer is employed, we have (14):

$$W'_{k_{i_1 j_1}} = W_{k_{i_1 j_1}} - \alpha \Delta W_{k_{i_1 j_1}} \quad (14)$$
$$W'_{k_{i_1 j_2}} = W_{k_{i_1 j_2}} - \alpha \Delta W_{k_{i_1 j_2}}$$

where $\alpha$ denotes the learning rates. Then we have (15):

$$W'_{k_{i_1 j_1}} - W'_{k_{i_1 j_2}} = \left( W_{k_{i_1 j_1}} - W_{k_{i_1 j_2}} \right)$$
$$- \alpha \left( \Delta W_{k_{i_1 j_1}} - \Delta W_{k_{i_1 j_2}} \right)$$
$$= \left( W_{k_{i_1 j_1}} - W_{k_{i_1 j_2}} \right)$$
$$- \alpha \left( \begin{array}{c} \left( \frac{\partial \mathcal{L}_T}{\partial W_{k_{i_1 j_1}}} - \frac{\partial \mathcal{L}_T}{\partial W_{k_{i_1 j_2}}} \right) + \\ \frac{2\lambda}{n_k} \left( \left( \sum_{i=1}^{n_k} P_{kij_1} - 1 \right) \Delta_{ij_1} - \left( \sum_{i=1}^{n_k} P_{kij_2} - 1 \right) \Delta_{ij_2} \right) \end{array} \right) \quad (15)$$

Based on (12) and (13), we have (16):

$$\left( \sum_{i=1}^{n_k} P_{kij_1} - 1 \right) \Delta_{kij_1} - \left( \sum_{i=1}^{n_k} P_{kij_2} - 1 \right) \Delta_{kij_2} > 0 \quad (16)$$

Thus, if we set $\lambda > \frac{\left(\frac{\partial \mathcal{L}_T}{\partial W_{k_{i_1 j_2}}} - \frac{\partial \mathcal{L}_T}{\partial W_{k_{i_1 j_1}}}\right) n_k}{2\left(\left(\sum_{i=1}^{n_k} P_{k_{ij_1}} - 1\right)\Delta_{k_{ij_1}} - \left(\sum_{i=1}^{n_k} P_{k_{ij_2}} - 1\right)\Delta_{k_{ij_2}}\right)}$, then (17) holds:

$$W'_{k_{i_1 j_1}} - W'_{k_{i_1 j_2}} - \left(W_{k_{i_1 j_1}} - W_{k_{i_1 j_2}}\right) < 0 \quad (17)$$

Therefore, by appropriately choosing $\lambda$, the gap between $W_{k_{i_1 j_1}}$ and $W_{k_{i_1 j_2}}$ will be gradually filled and after sufficient training epochs, $W_{k_{i_1 j_2}}$ will be greater than $W_{k_{i_1 j_1}}$. Then, according to **Proposition 1**, $P_{k_{i_1 j_1}} = 0$ and $P_{k_{i_1 j_2}} = 1$. Thus, the gap between $\sum_i P_{k_{ij_1}}$ and $\sum_i P_{k_{ij_2}}$ also will be gradually filled and finally $\sum_i P_{k_{ij_1}} = 1 = \sum_i P_{k_{ij_2}}$. This establishes **Proposition 2.**

## 4 COMPUTATIONAL EXPERIMENTS

We verify the effectiveness of the proposed RCO via computational experiments conducted on two widely studied data-driven modeling tasks in the renewable energy field, the wind speed prediction (WSP) and wind power prediction (WPP), which are able to consider multi-variate time-series data of numerous distributed units. The WSP and WPP are considered as two representative tasks because: 1) wind turbines as modern renewable energy systems are fully instrumented and commercial wind farms are commonly equipped with the supervisory control and data acquisition (SCADA) systems so that a large volume of SCADA data of wind turbines has been collected and available over past years; 2) a wind turbine itself is a complicated system having many attributes and its true dynamics is difficult to be modeled from a parametric aspect; and 3) there exists internal spatial-temporal interactions among wind speeds and power measured at different wind turbines due to many factors, such as the wake effect, wind kinetics, etc. Meanwhile, the WSP and WPP have strong practical implications since they are direct means for addressing the uncertainty of the wind power supply in a grid having a high wind power penetration due to the worldwide pursuit of the carbon neutrality.

### 4.1 Task description

Recently, to effectively capture the spatial-temporal data patterns, CNN has been wildly utilized in the WSP and WPP [5-9]. We adopt a CNN-LSTM structure studied in [6] as the base of modeling. The CNN was employed to capture the spatial patterns between WTs and the dynamics between attributes of WTs, and then output latent representations. Next, latent representations were ordered based on temporal relationships and fed into a LSTM to model the temporal correlation. Finally, a fully connected layer was applied to output the final WSP/WPP results. In [6-9], the input of the CNN was a two-dimensional tensor which was ordered manually. In this study, the RCO is utilized to rearrange the tensor input for the CNN. The objective function of data-driven WSP/WPP modeling problem is described as (18):

$$\min_\theta \mathcal{L}_T([X_0^1, X_0^2, ..., X_0^t], label) \\ = \min_\theta \|f([X_0^1, X_0^2, ..., X_0^t]), label\|_2 \quad (18)$$

where $X_0^1 \in R^{WT \times Attribute}$ denotes the input tensor of the CNN at timestamp 1, $t$ denotes the considered historical data length, $label \in R^s$ denotes the ground truth wind speed/power of the target WT, and $s$ denotes the considered prediction horizon. Four datasets sampled in the 10-min interval are utilized to conduct computational experiments and each dataset can be employed for both WPP and WSP. The detailed information of four datasets is provided in Table I.

TABLE I.
DATASETS STATISTICS

| Dataset | Number of turbines | Number of data points |
|---|---|---|
| Dataset 1 | 4 | 54283 |
| Dataset 2 | 25 | 26496 |
| Dataset 3 | 33 | 26350 |
| Dataset4 | 33 | 26350 |
| Considered parameters (WSP) | Wind speed, Wind Direction, Temperature | |
| Considered parameters of Dataset 1 (WPP) | Wind Power, Wind Speed, Wind Direction, Temperature, Generator Speed, Generator Temperature, Nacelle Angle, Nacelle Temperature, Pitch Angle | |
| Considered parameters of Dataset 2-4 (WPP) | Wind Power, Wind Speed, Nacelle Position, Wind Direction, Blade Angle (1), Blade Angle (2), Blade Angle (3), Generator Speed, Temperature, Nacelle Temperature, Vibration-X, Vibration-Y, Generator Temperature | |

# Dataset 1 is accessible via https://opendata-renewables.engie.com.

### 4.2 Parameter setup and evaluation metric

The CNN-LSTM model is adopted as the backbone of the CNN-based structure. The CNN contains two successive convolutional layers with the ReLU activation. The kernel size of two convolutional layers is $3 \times 3$. The channel of two convolutional layers is 32 and 64 respectively. The number of hidden neurons of LSTM is set to 128. The $\lambda$ in Eq. (7) is set to 1. The t and s in Eq. (18) are set to 50 and 6, respectively. Each experiment is repeated three times with random seeds. The Root Mean Square Error (RMSE) expressed in (19) is utilized for the performance evaluation.

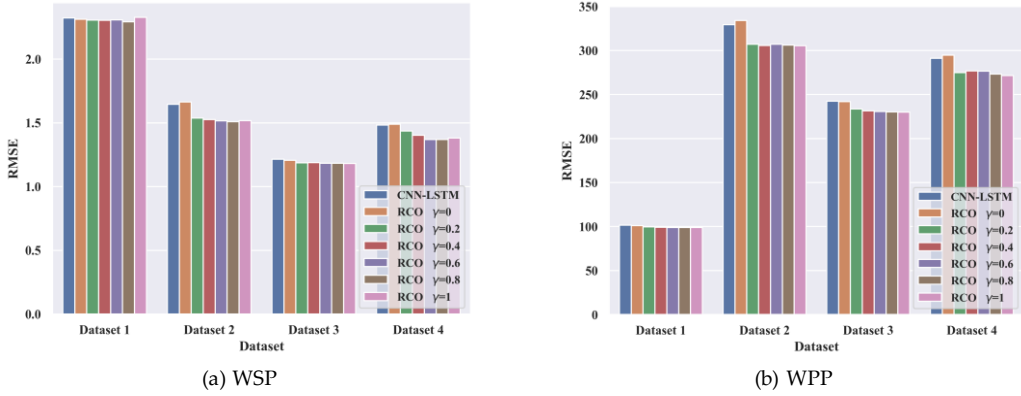

(a) WSP  (b) WPP

Fig. 5. The bar plot of the prediction with different $\gamma$.

TABLE II. THE RESULTS OF WPP

| Dataset | CNN-LSTM | CNN-LSTM with RCO | | | | | |
|---|---|---|---|---|---|---|---|
| | | $\gamma = 0$ | $\gamma = 0.2$ | $\gamma = 0.4$ | $\gamma = 0.6$ | $\gamma = 0.8$ | $\gamma = 1$ |
| Dataset 1 | 101.44±0.59 | 101.11±0.46 | 99.61±0.32 | 99.02±0.23 | 98.84±0.27 | **98.76±00.21** | 98.85±0.22 |
| Dataset 2 | 329.27±1.60 | 334.16±2.66 | 307.20±0.80 | 305.81±1.33 | 307.03±1.10 | 306.18±0.27 | **305.35±1.45** |
| Dataset 3 | 242.32±1.29 | 242.02±1.26 | 233.52±0.74 | 231.40±0.86 | 230.71±0.55 | 230.25±0.77 | **230.03±0.08** |
| Dataset 4 | 291.11±1.64 | 294.78±2.58 | 274.93±4.25 | 276.75±1.86 | 276.64±2.11 | 273.13±2.02 | **271.46±2.27** |
| Average | 276.21 | 278.66 | 261.65 | 261.16 | 261.19 | 259.59 | **258.71** |

TABLE III. THE RESULTS OF WSP

| Dataset | CNN-LSTM | CNN-LSTM with RCO | | | | | |
|---|---|---|---|---|---|---|---|
| | | $\gamma = 0$ | $\gamma = 0.2$ | $\gamma = 0.4$ | $\gamma = 0.6$ | $\gamma = 0.8$ | $\gamma = 1$ |
| Dataset 1 | 2.323±0.015 | 2.313±0.016 | 2.306±0.004 | 2.306±0.017 | 2.308±0.014 | **2.293±0.003** | 2.329±0.035 |
| Dataset 2 | 1.646±0.022 | 1.665±0.006 | 1.538±0.005 | 1.526±0.004 | 1.515±0.001 | **1.510±0.001** | 1.517±0.002 |
| Dataset 3 | 1.215±0.001 | 1.206±0.004 | 1.187±0.002 | 1.189±0.002 | 1.183±0.003 | 1.184±0.002 | **1.182±0.001** |
| Dataset 4 | 1.482±0.007 | 1.490±0.028 | 1.436±0.028 | 1.404±0.021 | 1.370±0.015 | **1.369±0.006** | 1.382±0.008 |
| Average | 1.667 | 1.669 | 1.617 | 1.606 | 1.594 | **1.589** | 1.602 |

$$RMSE = \sqrt{\frac{1}{n}\sum_{i=1}^{n}\|label - prediction\|_2^2} \quad (19)$$

## 4.3 Computational results and analyses

Fig. 5 employs bar charts to present the average *RMSE* of the CNN-LSTM and the CNN-LETM with the proposed RCO under different $\gamma$. It is observable that, given $\gamma = 0$, which means the generic permutation matrix is adopted, the performance improvement is negligible. By increasing $\gamma$ from 0 to 1 gradually, the accuracy of WPP based on the CNN-LSTM with RCO improves gradually. However, given $\gamma = 1$, which means that we do not regularize the feature diversity, the performance may decline in a few scenarios. Average prediction results in WSP and WPP are reported in Tables II and III. It is observable that, for the WPP task, by setting $\gamma = 1$, the proposed RCO could significantly improve the accuracy of the original CNN-LSTM. While in WSP, a better performance can be achieved by setting $\gamma = 0.8$.

Figs. 6-9 visualize the permutation matrices of one WT randomly selected from Dataset 1 and Dataset 2 for the WPP task, where the wind power output is the first attribute of the original data, and $P_1$ and $P_2$ are the permutation matrices for the first and the second axes of $X_0 \in R^{WT \times Attribute}$ respectively. It is observable that by adopting the Gumbel-Softmax and the proposed soft regularization loss, the permutation matrix could finally converge and meet both binary and right stochastic conditions. Given $\gamma = 0$, the learned permutation matrices are doubly stochastic matrices and, by increasing $\gamma$, local correlations appear repeatedly, which confirms that the local correlation and the shift

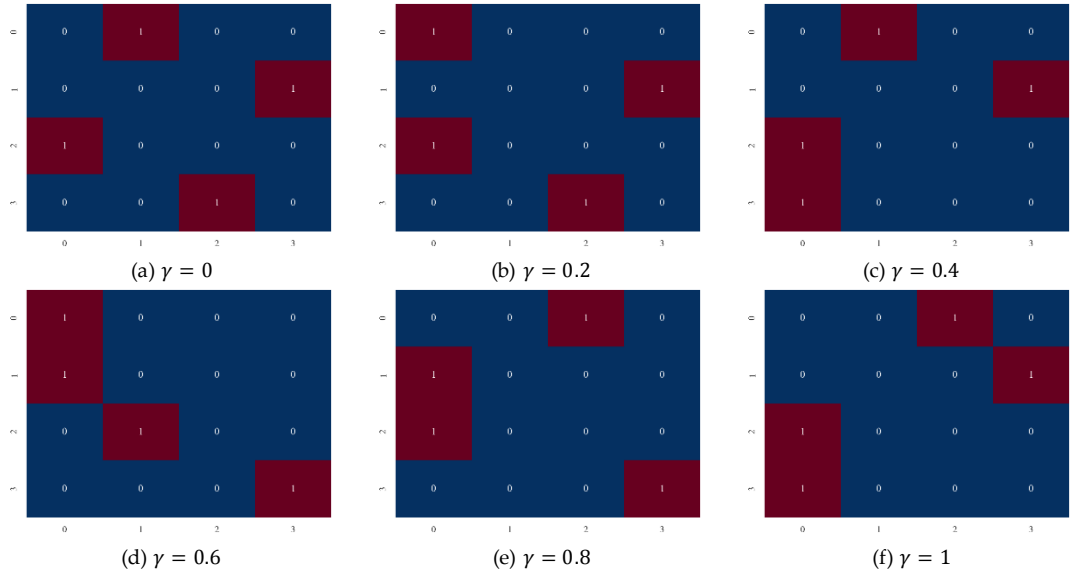
Fig. 6. An illustration of $P_1$ of Dataset 1 for WPP.

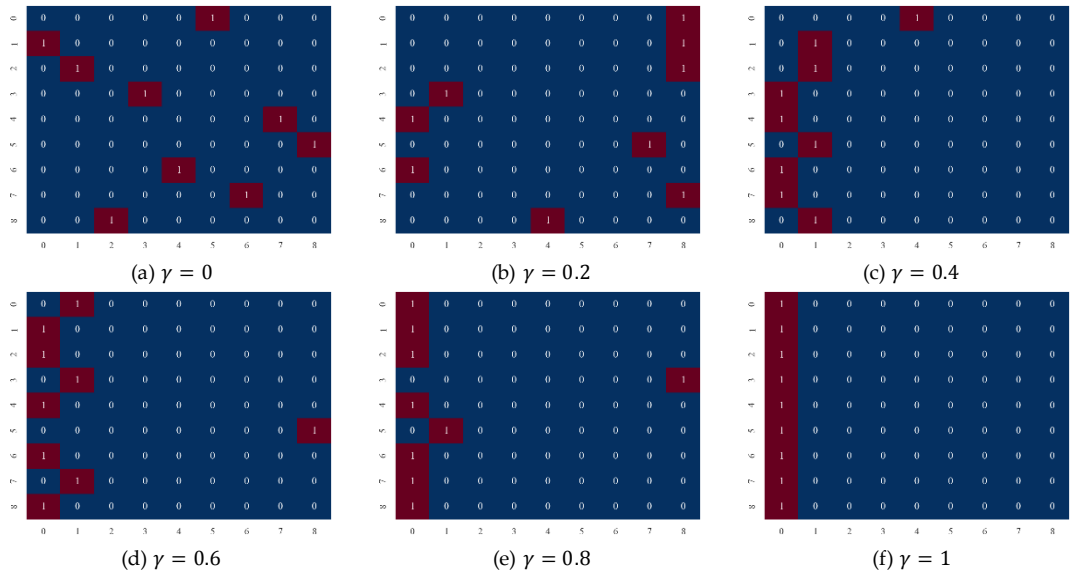
Fig. 7. An illustration of $P_2$ of Dataset 1 for WPP.

TABLE. IV. THE AVERAGE COMPUTATIONAL TIME

| Method | Training Time (s) | Testing Time (s) |
|---|---|---|
| CNN-LSTM | 7.283 | 3.260e-5 |
| CNN-LSTM with RCO | 9.425 | 3.279e-5 |

invariance properties of the industrial data based tensor after re-arranging are enhanced. However, given $\gamma = 1$, the permutation matrix may only pay attention to the most important local features, e.g., the wind power parameter, which may result in overfitting and impair the performance.

Fig. 10 visualizes a set of randomly chosen periods of the WPP and WSP results. It is clear that the proposed RCO could improve the performance of the original CNN-LSTM and offer better predictions.

Table IV reports the averaged training/testing time of the CNN-LSTM and the CNN-LSTM with the proposed RCO over one epoch. It is clear that training CNN-LSTM with the proposed RCO consumes slightly more time by comparing with that of the original CNN-LSTM. While, the CNN-LSTM with the proposed RCO has a comparable testing time to the original CNN-LSTM. These results verify the effectiveness and efficiency of the proposed RCO in the real implementation.

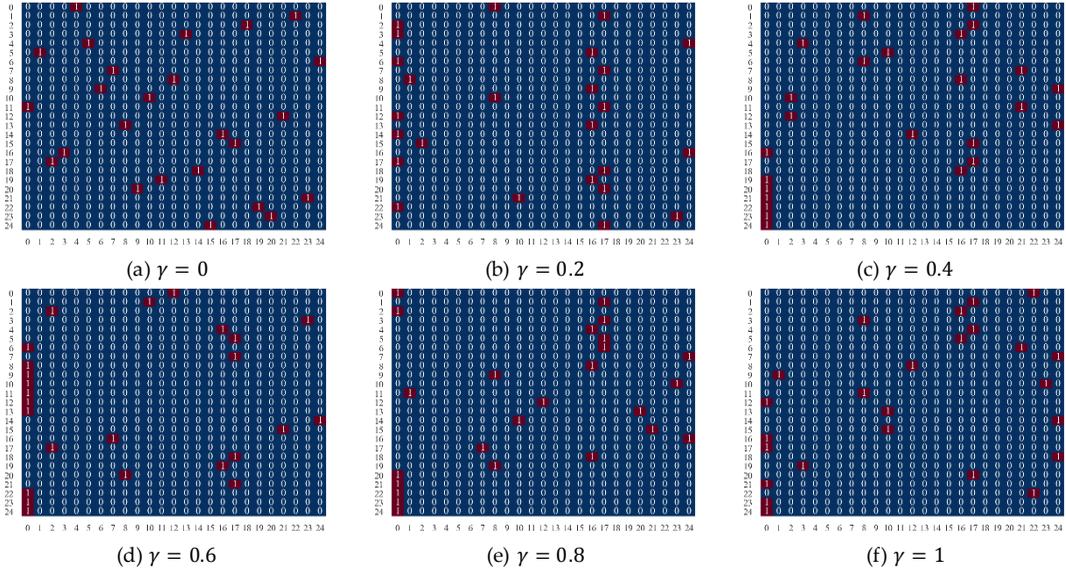

Fig. 8. An illustration of $P_1$ of Dataset 2 for WPP.

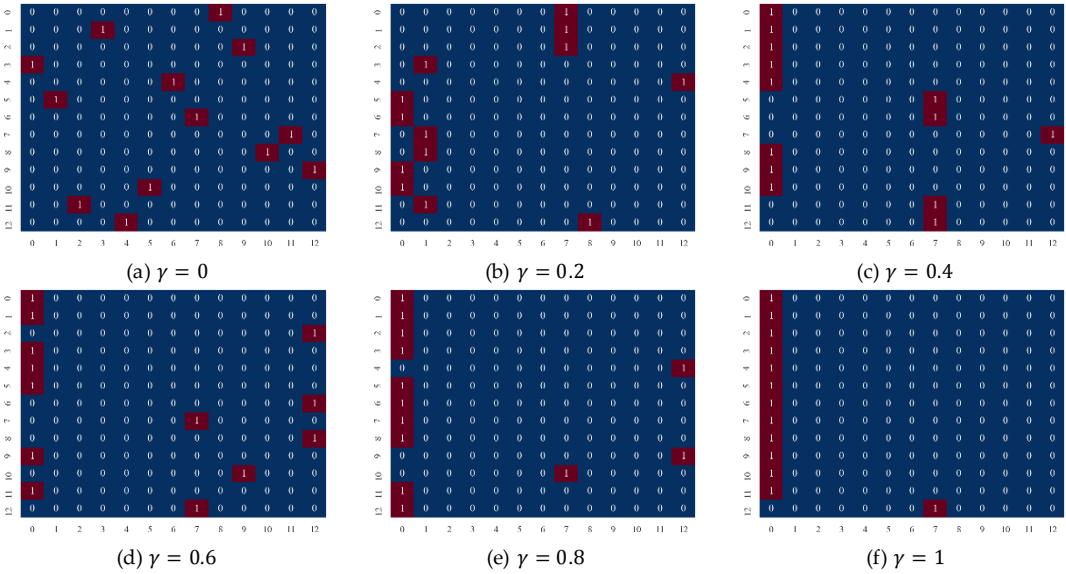

Fig. 9. An illustration of $P_2$ of Dataset 2 for WPP.

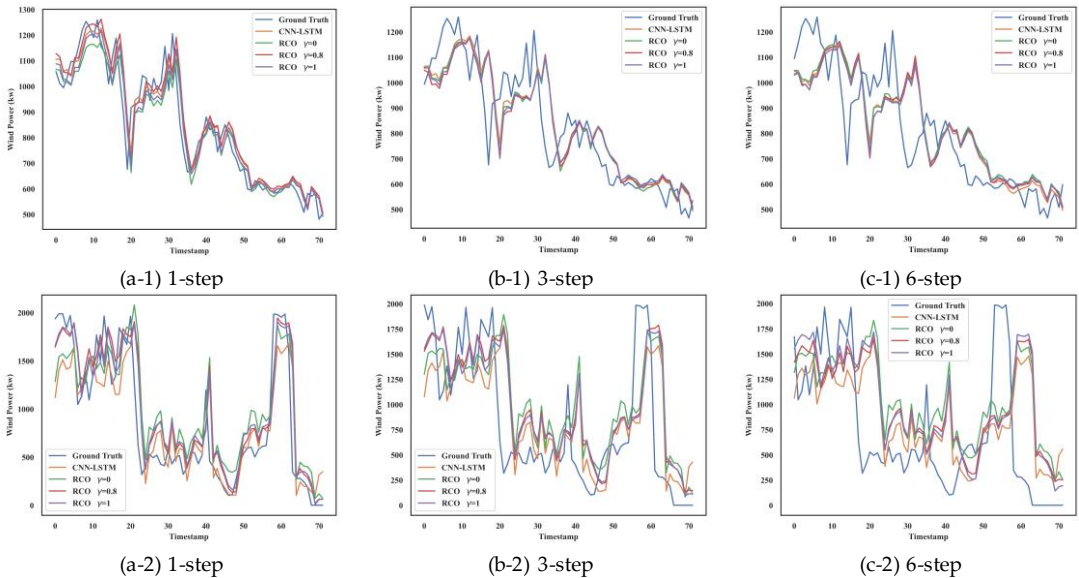

Fig. 10. An illustration of WPP results.

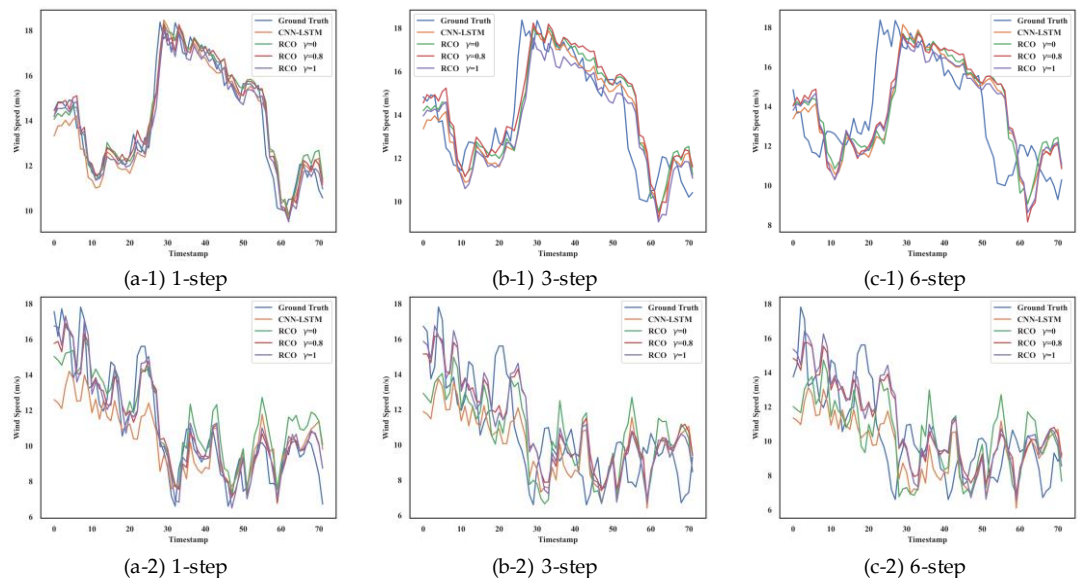

Fig. 11. An illustration of WSP results.

## 5 CONCLUSION

In this paper, a plug and play module, RCO, was proposed to automatically rearrange the order of attributes of the industrial data based tensor before the downstream CNN based learning tasks. The proposed RCO maintained $K$ permutation matrices to permutate $K$ axes of the input industrial data based tensor. To satisfy the binary condition and the right stochastic condition, a novel learning process was proposed, where the Gumbel-Softmax was utilized to reparameterize the elements of the permutation matrices, and the soft regulation loss was proposed to preserve the feature diversity of the industrial data based tensor. The proposed RCO could be updated together with the subsequent CNNs via the gradient-based algorithm and was verified based on computational experiments considering two famous tasks, the WPP and WSP, in the renewable energy domain. The computational results demonstrated that the proposed RCO could significantly improve the performance of the generic CNN-based network.